\documentclass[sigconf]{acmart}

\usepackage{subcaption}

\setcopyright{none}

\renewcommand\footnotetextcopyrightpermission[1]{}
\acmConference{\relax}{\relax}{\relax}

\begin{document}

\title[Enhancing Online Learning Efficiency...]{Enhancing Online Learning Efficiency Through Heterogeneous Resource Integration with a Multi-Agent RAG System}

\author{Devansh Srivastav}
\affiliation{
    \institution{German Research Center for Artificial Intelligence (DFKI)}
    \city{Saarbrücken}
    \country{Germany}
}
\affiliation{
    \institution{Saarland University}
    \city{Saarbrücken}
    \country{Germany}
}
\email{devansh.srivastav@dfki.de}

\author{Hasan Md Tusfiqur Alam}
\affiliation{
    \institution{German Research Center for Artificial Intelligence (DFKI)}
    \city{Saarbrücken}
    \country{Germany}
}
\email{hasan.alam@dfki.de}

\author{Afsaneh Asaei}
\affiliation{
    \institution{Digital Product School of UnternehmerTUM (Center for Innovation and Business Creation at Technical University of Munich)}
    \city{Munich}
    \country{Germany}
}
\email{asaei@unternehmertum.de}

\author{Mahmoud Fazeli}
\affiliation{
    \institution{CROWDCONSULTANTS}
    \city{Berlin}
    \country{Germany}
}
\email{mahmoud.fazeli@crowdconsultants.com}

\author{Tanisha Sharma}
\affiliation{
    \institution{Microsoft}
    \city{Hyderabad}
    \country{India}
}
\email{tansharma@microsoft.com}

\author{Daniel Sonntag}
\affiliation{
    \institution{German Research Center for Artificial Intelligence (DFKI)}
    \city{Saarbrücken}
    \country{Germany}
}
\affiliation{
    \institution{University of Oldenburg}
    \city{Oldenburg}
    \country{Germany}
}
\email{daniel.sonntag@dfki.de}

\renewcommand{\shortauthors}{Srivastav et al.}

\begin{abstract}
Efficient online learning requires seamless access to diverse resources such as videos, code repositories, documentation, and general web content. This poster paper introduces early-stage work on a Multi-Agent Retrieval-Augmented Generation (RAG) System designed to enhance learning efficiency by integrating these heterogeneous resources. Using specialized agents tailored for specific resource types (e.g., YouTube tutorials, GitHub repositories, documentation websites, and search engines), the system automates the retrieval and synthesis of relevant information. By streamlining the process of finding and combining knowledge, this approach reduces manual effort and enhances the learning experience. A preliminary user study confirmed the system's strong usability and moderate-high utility, demonstrating its potential to improve the efficiency of knowledge acquisition.
\end{abstract}



\maketitle

\section{Introduction}
In an increasingly interconnected world, the ability to access, integrate, and utilize diverse online resources is essential for effective learning and knowledge acquisition \cite{lebenicnik2015use,tammets2022integrating}. With the rapid growth of digital platforms, learners often rely on a combination of heterogeneous resources, such as video tutorials, code repositories, documentation websites, and search engines, to acquire new skills and understanding concepts \cite{alghamdi2023novice,iftikhar2019impact}. However, navigating and synthesizing information across these disparate sources can be a time-intensive and inefficient process, creating barriers to efficient online learning \cite{skulmowski2022understanding}.
The challenges associated with multi-source learning are especially evident in technical domains, where the need to quickly find accurate and relevant information is critical. For instance, a developer exploring a new framework might consult a YouTube tutorial for an overview, reference a GitHub repository for implementation details, examine the official documentation for deeper insights, and conduct general web searches for troubleshooting. Manually searching, cross-referencing, and consolidating information from these varied sources often leads to cognitive overload and suboptimal learning experiences.

To address these challenges, this poster paper proposes a Multi-Agent RAG system that automates the retrieval and synthesis of information from diverse online resources. By leveraging specialized agents tailored for specific resource types—such as YouTube tutorials, GitHub repositories, documentation websites, and general web searches—the system streamlines the process of finding, integrating, and utilizing knowledge. This approach reduces the manual effort required for cross-referencing and consolidating information, enabling users to focus on learning and problem-solving rather than on navigating disparate sources. The system aims to enhance the learning experience by providing a unified, user-friendly interface for targeted, real-time inquiries, making it a valuable tool for learners and developers in technical domains.

\begin{figure}[ht!]
    \centering
    \includegraphics[width=\linewidth]{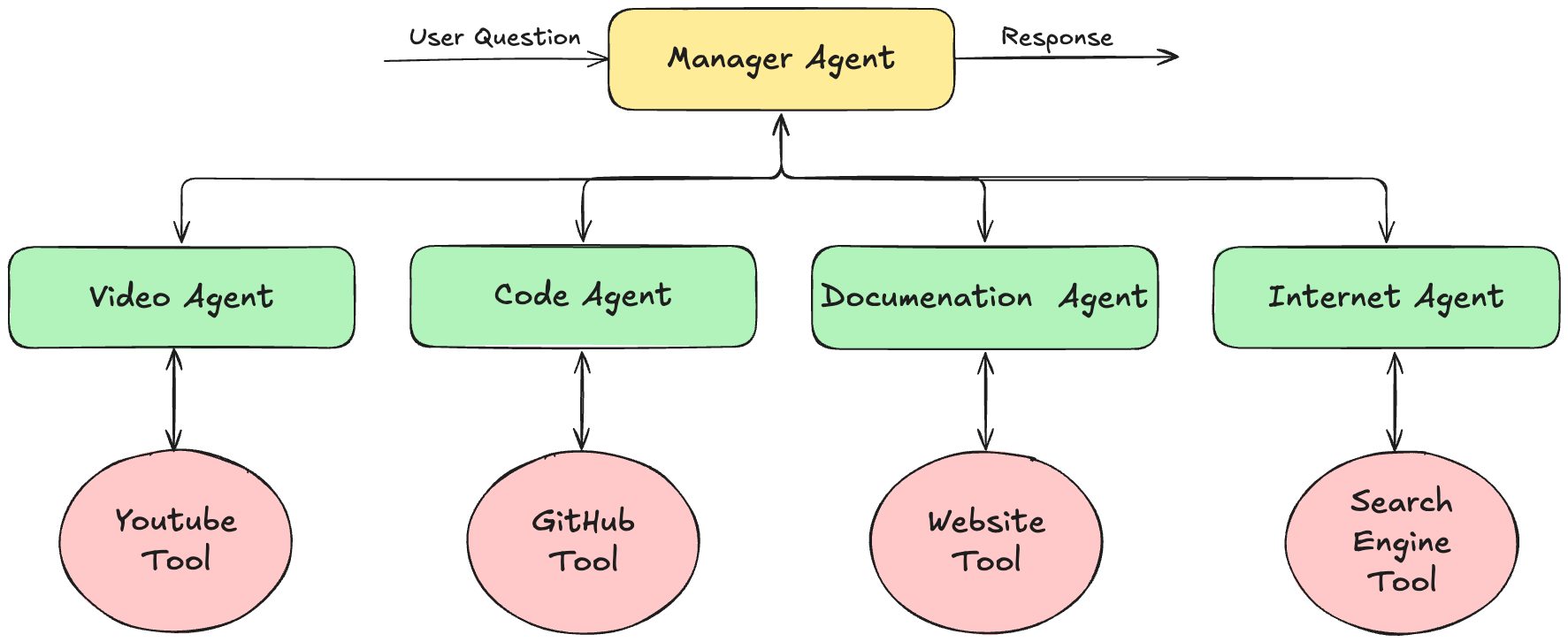}
    \caption{Working Flow of the Multi-Agent RAG System}
    \Description{The diagram illustrates the working flow of the Multi-Agent RAG System, highlighting how user queries are processed and responses are generated. At the center of the system is the Manager Agent, which acts as a coordinator, receiving questions from the user and delegating tasks to four specialized agents. These agents include the Video Agent, connected to a YouTube Tool; the Code Agent, connected to a GitHub Tool; the Documentation Agent, connected to a Website Tool; and the Internet Agent, connected to a Search Engine Tool. Each agent operates independently, retrieving information relevant to its specific resource type. Once the agents have completed their retrieval tasks, the Manager Agent synthesizes the outputs from all agents into a cohesive response, which is then delivered back to the user. The diagram visually represents this process using arrows to depict the flow of tasks and information between the Manager Agent, the specialized agents, and their corresponding tools.}
    \label{fig:arch}
\end{figure}

\section{System Architecture and Implementation}
The proposed system uses a Multi-Agent RAG system with GPT-4o \cite{hurst2024gpt} as the underlying LLM. The flow, as shown in Fig. \ref{fig:arch}, begins with a central Manager Agent that coordinates task delegation across four specialized agents, each tailored to handle a specific type of resource: YouTube Video, GitHub Repository, Documentation Website, and a Generic Search Engine. These agents operate independently, leveraging resource-specific tools and APIs to retrieve relevant information. The Manager Agent synthesizes outputs from these agents, combining them into a cohesive response to address the user’s query. This modular design ensures scalability and adaptability, allowing for seamless integration of additional agents or tools as needed to meet evolving requirements.

\begin{figure}[b!]
    \centering
    \begin{subfigure}{0.85\linewidth}
        \centering
        \includegraphics[width=0.85\linewidth]{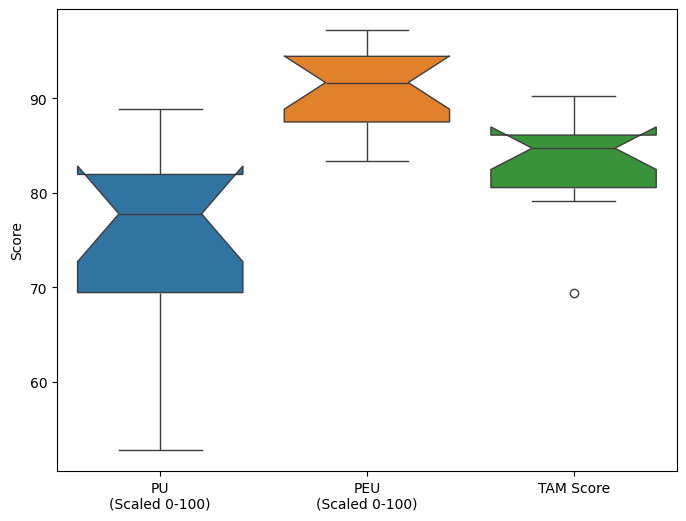}
        \caption{PU and PEU score distribution}
        \label{fig:box}
    \end{subfigure}
    \begin{subfigure}{0.85\linewidth}
        \centering
        \includegraphics[width=0.85\linewidth]{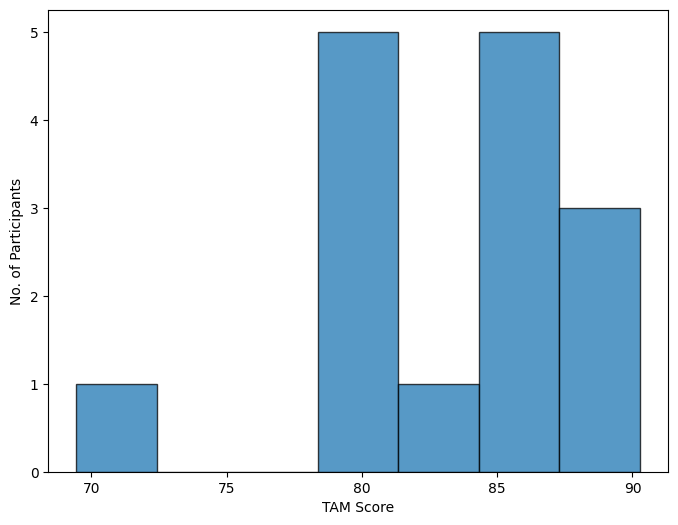}
        \caption{TAM score distribution for 15 participants.}
        \label{fig:hist}
    \end{subfigure}
    \caption{Analysis of the user study using TAM}
    \label{fig:plots}
    \Description{The plots visually represent the results from the preliminary evaluation of the Multi-Agent RAG system, as described in the study. The left panel shows notched box plots for Perceived Usefulness (PU), Perceived Ease of Use (PEU), and the overall TAM (Technology Acceptance Model) scores, all scaled from 0 to 100. The PU scores exhibit a wider spread, ranging from about 60 to 90, indicating variability in participants' perceptions of the system's utility. The PEU scores, in contrast, show a tighter distribution, with most values close to the upper end of the scale (around 90), reflecting strong consensus on the system's usability. The TAM scores are also tightly distributed but slightly lower than the PEU scores, reflecting the influence of both constructs. The right panel presents a histogram of TAM scores, showing the distribution of overall scores among participants. Most participants scored the system between 80 and 90, with peaks at 85 and 90, indicating a generally favorable reception. A few participants scored lower, around 70, suggesting some variability in overall satisfaction, likely tied to differences in perceptions of usefulness. Together, these plots underscore the system’s strong usability and moderate utility, as highlighted in the preliminary evaluation section.}
\end{figure}

\begin{figure*}[h]
    \centering
    \vspace{0.6em}
    \includegraphics[width=0.70\linewidth, trim={0 0.2cm 3cm 4cm},clip]
    {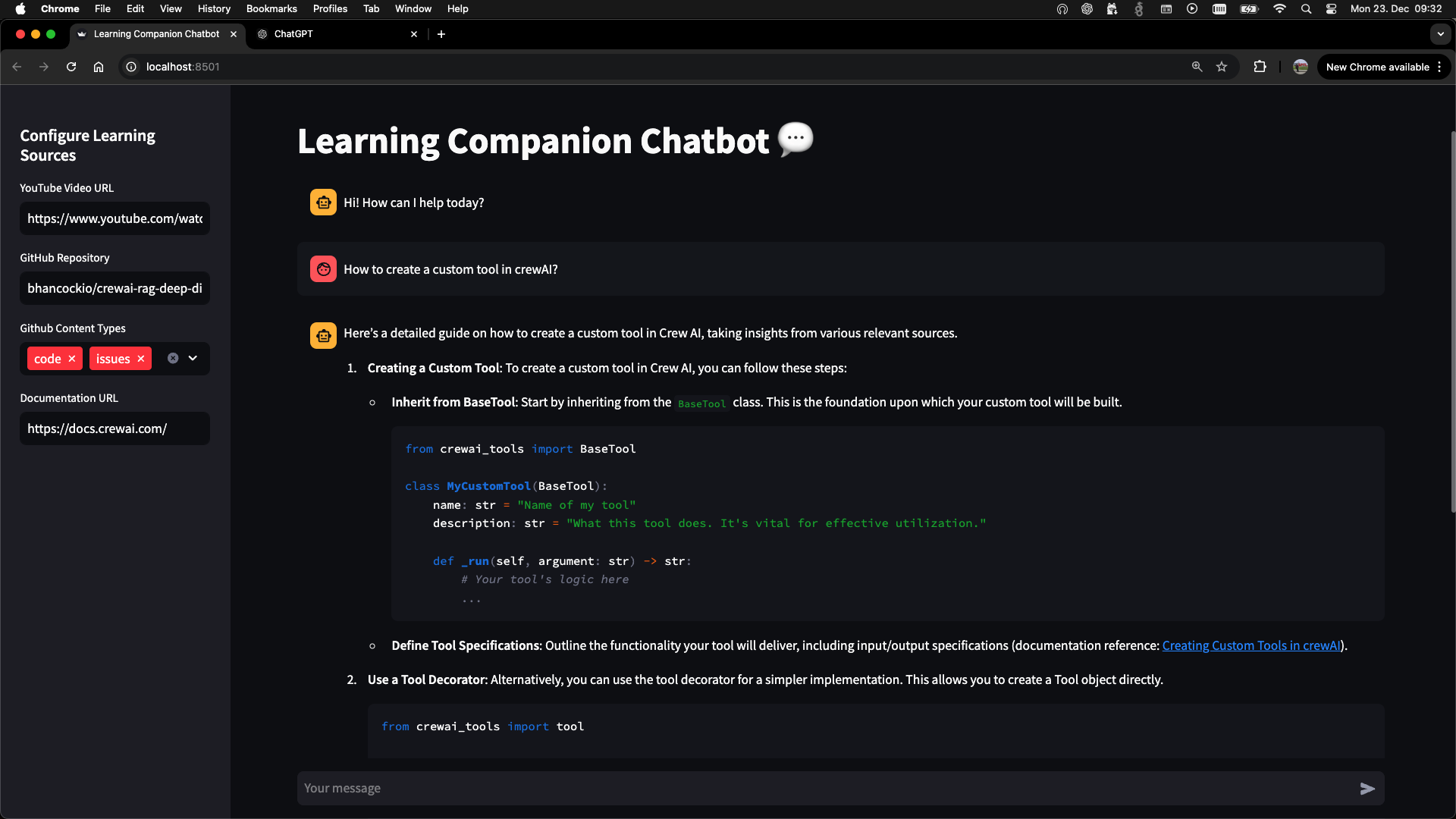}
    \includegraphics[width=0.70\linewidth, trim={0 0.2cm 3cm 4cm},clip]{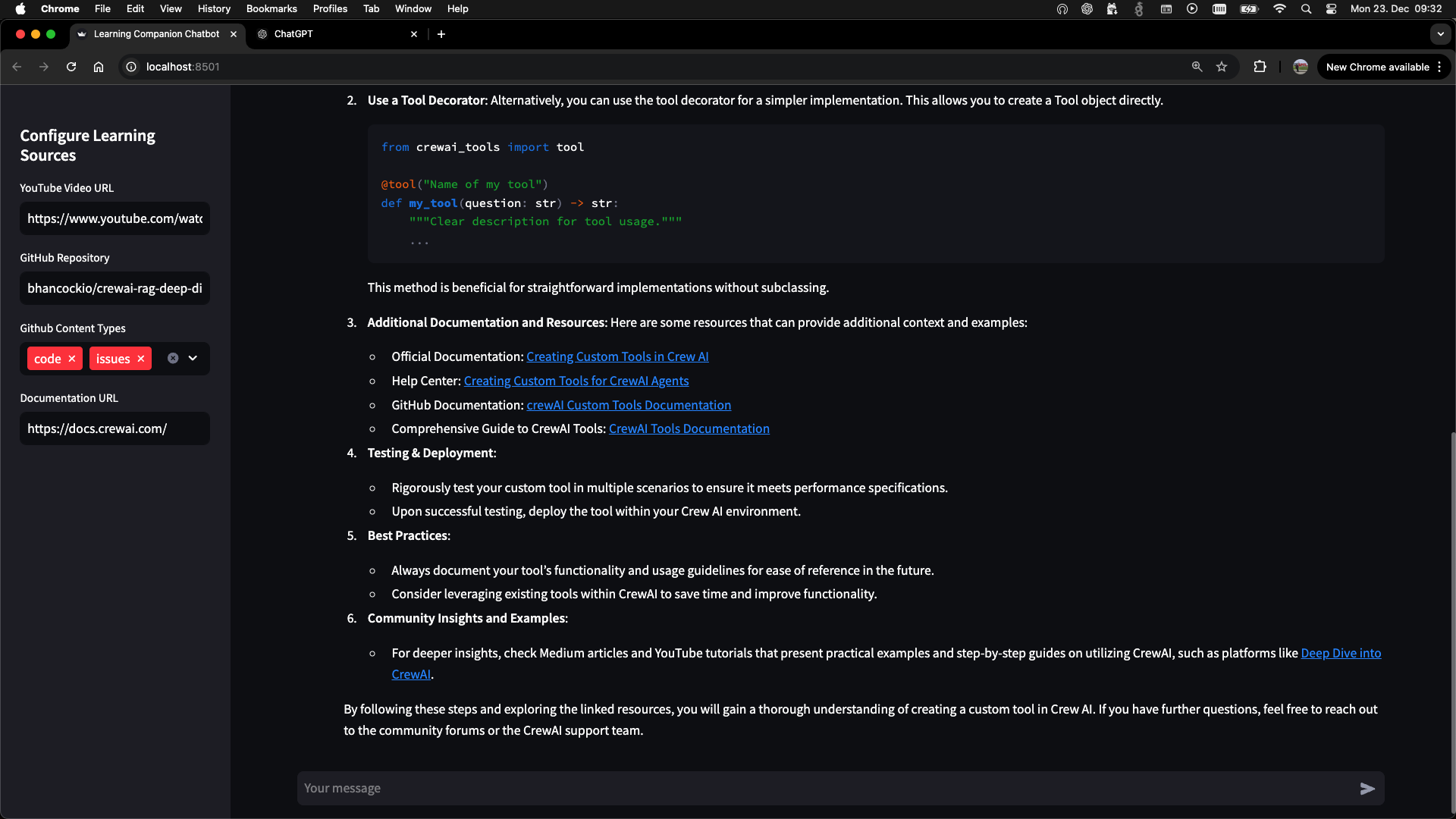}
    \Description{The figure showcases the user interface of the Learning Companion Chatbot, implemented using Streamlit, designed for configuring and querying learning resources. The interface is divided into two main sections. On the left, users can input and configure the sources of learning material, such as a YouTube video URL, GitHub repository URL, documentation URL, and specific GitHub content types like code or issues. This configuration panel allows for tailored resource integration based on user preferences. On the right, the chatbot provides a conversational interface where users can interact with the system by asking targeted questions. In the example displayed, a user queries, “How to create a custom tool in CrewAI?” The chatbot responds with detailed, step-by-step instructions for creating a custom tool, including code snippets and explanations. The response is well-structured, with sections for defining tool specifications, using decorators, and exploring additional resources such as documentation and community forums. This interface demonstrates the system's capability to streamline learning by enabling users to configure sources effectively and receive contextualized, actionable insights in real time. The clean layout and intuitive design make it accessible for both technical and non-technical users.}
    \caption{User Interface of the proposed system.}
    \label{fig:screenshot}
\end{figure*}

Data from sources like YouTube transcripts, GitHub repositories, and documentation websites is preprocessed to extract relevant text. Embeddings are generated using GPT-4o, to capture the semantic structure of the text. These embedding vectors, along with metadata such as source URLs and timestamps, are stored in ChromaDB, enabling precise semantic searches and contextually relevant results. For general web content, the system retrieves information dynamically based on the user’s query.
The user interface (UI) for the system is implemented using Streamlit \cite{khorasani2022web}, and it allows users to specify the URLs of the YouTube tutorial, GitHub repository, and documentation webpage as inputs. Additionally, the types of GitHub content to include such as code, issues, or pull requests, can be specified, allowing the system's focus to be tailored to specific needs. Users can then conversationally interact with the system, asking targeted questions about the provided resources as shown in Fig. \ref{fig:screenshot}. This functionality empowers users to focus their queries on preselected sources, reducing irrelevant results and improving the relevance and accuracy of responses. The real-time interaction capabilities of the UI ensure that users receive immediate feedback on their questions, enhancing the overall user experience. The simplicity and intuitiveness of the system make it accessible to users with varying levels of technical expertise, lowering the barrier to entry for non-technical audiences. 

\section{Preliminary Evaluation}
To evaluate the system’s usability and utility, a user study was conducted involving 15 participants aged 21 to 29 (mean age = 24 years; SD = 2.26), all of whom were master's-level students pursuing computer science-related courses. These participants were selected due to their familiarity with the process of learning new programming languages or frameworks using various online resources such as YouTube tutorials, GitHub repositories, and documentation websites. This familiarity enabled them to assess whether the proposed system adds value to their learning process by streamlining the integration and retrieval of information from these resources. The study used the Technology Acceptance Model (TAM) \cite{davis1989technology} questionnaire to assess technology acceptance through two key constructs: Perceived Usefulness (PU) and Perceived Ease of Use (PEU). The box plot of PU, PEU, and TAM Scores in Fig. \ref{fig:box} reveals considerable variability in PU ratings compared to PEU. The average PU score, scaled to a 0–100 range \cite{lewis2019comparison}, was 75, indicating moderate-high utility. The wider spread of PU scores indicates that while some participants perceive the system as highly useful, others find its utility less apparent highlighting that the system may not fully align with the user's needs or expectations. This variability suggests the potential for targeted improvements to better meet the requirements of all users. In contrast, the average PEU score was 91.11, with the box plot showing a tighter distribution of scores. This consistency suggests that participants broadly agree on the system’s strong usability. The TAM score histogram in Fig. \ref{fig:hist} revealed most users scored above 80 with the overall TAM scores, averaging 83.06. This reflects the combined influence of the PU and PEU constructs. While the high PEU scores contribute significantly to the favourable TAM scores, the variability in PU remains a critical factor for consideration.

\section{Conclusion and Future Work}
The proposed Multi-Agent RAG system automates the retrieval and synthesis of relevant information, effectively integrating resources such as YouTube tutorials, GitHub repositories, documentation websites, and general web content. By streamlining the process of finding and combining knowledge, the system reduces manual effort and enhances the learning experience. Evaluation results highlight strong usability, with high perceived ease of use reflecting the system’s accessible and user-friendly interface. While the system demonstrates considerable utility, variability in perceived usefulness indicates a need for better alignment with user expectations and task-specific needs. 

Future work will focus on making the retrieval system dynamic, enabling it to determine when specialized agents are necessary to improve response times and reduce computational overhead. It will also involve testing with participants less familiar with technical environments to evaluate the system's acceptance across varying expertise levels, expanding resource integration to include domain-specific platforms like Stack Overflow, broadening its applicability to both technical and non-technical domains, and personalization to tailor responses to individual preferences, enhancing usability and user satisfaction. 


\bibliographystyle{ACM-Reference-Format}
\bibliography{main}

\end{document}